\documentclass{article} 
\usepackage{iclr2025_conference,times}

\usepackage{amsmath,amsfonts,bm}

\def\eqref#1{equation~\ref{#1}}

\def\1{\bm{1}}

\DeclareMathAlphabet{\mathsfit}{\encodingdefault}{\sfdefault}{m}{sl}
\SetMathAlphabet{\mathsfit}{bold}{\encodingdefault}{\sfdefault}{bx}{n}

\usepackage{microtype}
\usepackage{hyperref}
\usepackage{url}
\definecolor{darkblue}{rgb}{0, 0, 0.5}
\hypersetup{colorlinks=true, citecolor=darkblue, linkcolor=darkblue, urlcolor=darkblue}
\usepackage{geometry}
\usepackage{array}
\usepackage{enumitem}
\usepackage{graphicx}
\usepackage{subcaption}
\usepackage{booktabs}
\usepackage{arydshln}
\usepackage{multirow}
\usepackage{tabularx}
\usepackage{makecell}
\usepackage{amsmath}
\usepackage{amssymb}
\usepackage{amsfonts}
\usepackage{mathtools}
\usepackage{fontawesome}
\usepackage{todonotes}

\usepackage{xspace}

\usepackage{wrapfig}

\usepackage{tcolorbox}

\newcommand{\qwenname}{Qwen2.5\xspace}
\newcommand{\qwenbasemodel}{\textcolor{blue}{Qwen2.5-1.5B-Instruction}\xspace}
\newcommand{\qwenllm}{\textcolor{blue}{Qwen2.5-32B-Instruction}\xspace}
\newcommand{\fullname}{\href{https://huggingface.co/jinaai/ReaderLM-v2}{\texttt{ReaderLM-v2}}\xspace}
\newcommand{\shortname}{\href{https://huggingface.co/jinaai/ReaderLM-v2}{\texttt{ReaderLM-v2}}\xspace}
\title{\fullname: Small Language Model for HTML to Markdown and JSON}

\iclrfinalcopy
\author{Feng Wang, Zesheng Shi\thanks{work done while at Jina AI}~, Bo Wang, Nan Wang, Han Xiao\\
Jina AI GmbH\\
Prinzessinnenstr. 19-20, 10969\\
Berlin, Germany \\
\texttt{research@jina.ai}
}

\begin{document}

\maketitle

\begin{abstract}
We present \shortname, a compact 1.5 billion parameter language model designed for efficient web content extraction.
Our model processes documents up to 512K tokens, transforming messy HTML into clean Markdown or JSON formats with high accuracy—making it an ideal tool for grounding large language models.
The model's effectiveness results from two key innovations: (1) a three-stage data synthesis pipeline that generates high-quality, diverse training data by iteratively drafting, refining, and critiquing web content extraction;
and (2) a unified training framework combining continuous pre-training with multi-objective optimization.
Intensive evaluation demonstrates that \shortname outperforms GPT-4o-2024-08-06 and other larger models by 15-20\% on carefully curated benchmarks, particularly excelling at documents exceeding 100K tokens, while maintaining significantly lower computational requirements.
\end{abstract}



\begin{figure}[htbp]
    \centering
    \includegraphics[width=0.85\textwidth]{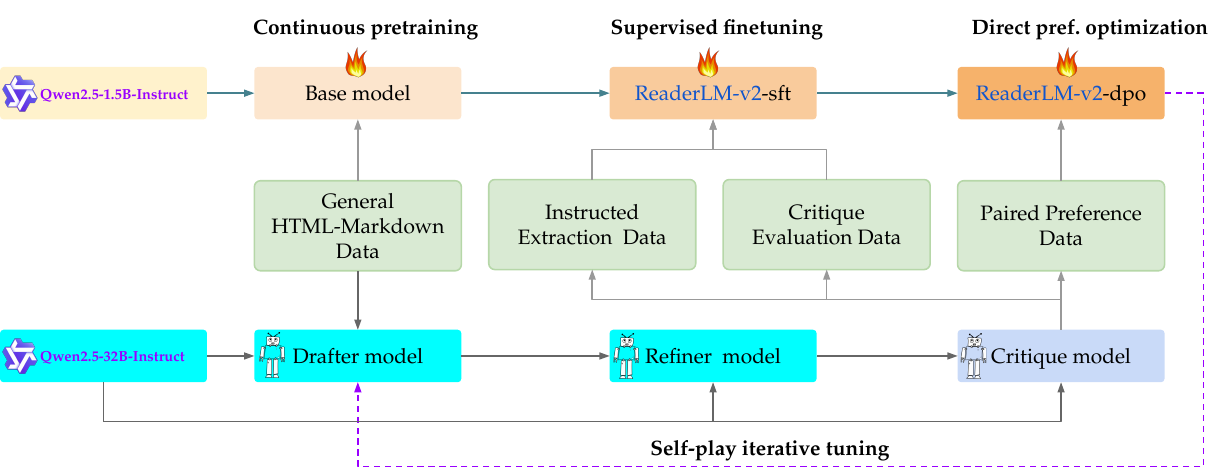}
    \caption{\shortname's iterative training process.
    Our approach combines (a) a novel three-stage data synthesis pipeline (\textsc{Draft-Refine-Critique}) that generates high-quality training data,
    with (b) a comprehensive training strategy incorporating continuous pre-training, supervised fine-tuning, direct preference optimization, and self-play iterative tuning.
    The iterative nature of both components allows for continuous model improvement through cycles of data generation and model refinement.}
    \label{fig:overview}
\end{figure}
\vfill

\newpage

\section{Introduction}
\label{sec:introduction}

The emergence of Large Language Models (LLMs) has revolutionized natural language processing tasks~\citep{brown2020language}.
One particularly useful application is structured content extraction,
which involves converting messy, unstructured data into clean, structured formats suitable for humans, machines, and LLMs, such as JSON or Markdown.
This extracted data is beneficial for tasks like retrieval, matching, and knowledge grounding, where well-structured data significantly enhances downstream performance. Many industries rely on these capabilities for applications like document processing, customer service automation, and knowledge management systems.

Despite these practical needs, LLM-based content extraction faces numerous challenges, including hallucination~\citep{ji2022survey}, limited context windows~\citep{press2022train}, computational inefficiency~\citep{tay2022efficient}, and the inability of many popular models to be hosted locally.
Addressing these challenges is crucial to enabling accurate and scalable structured content extraction.

In this research, we present \shortname, a specialized 1.5 billion parameter model fine-tuned on top of \qwenname~\citep{qwen2024qwen25},
designed to transform messy HTML data into JSON and Markdown formats.
Our findings demonstrate that a fine-tuned Small Language Model (SLM) can achieve on-par or even better performance compared to much larger or proprietary LLMs in the domain of structured content extraction.

The key contributions of our work are twofold:
First, we introduce a novel three-stage data synthesis pipeline, called \textbf{Draft-Refine-Critique}, which generates high-quality training data through iterative refinement; and second, we propose a comprehensive training strategy that combines continuous pre-training for length extension, supervised fine-tuning with specialized checkpoints, direct preference optimization (DPO), and self-play iterative tuning.
To facilitate further research and application of structured content extraction, the model is publicly available on Hugging Face\footnote{\url{https://huggingface.co/jinaai/ReaderLM-v2}}.

\section{Model Objectives}
\label{sec:task_definition}

Our aim is to create a model to transform raw HTML content into structured formats, such as JSON or Markdown. This is inherently complex due to several real-world challenges outlined below:

\begin{enumerate}
    \item \textbf{HTML Complexity:} Real-world HTML is inherently messy, often deviating from schema standards. It may include legacy markup, embedded JavaScript, comments, and CSS, which complicates extraction and parsing.

    \item \textbf{Length Variability:} As shown in Figure \ref{fig:html-length-distribution}, HTML documents exhibit a long-tail token length distribution. 
    Current LLMs, such as \qwenname which supports 128K context length through the YARN length extension method~\citep{peng2024yarn}, still struggle to process this variability effectively, often missing critical content in longer inputs.

    \item \textbf{Inference Bottlenecks:} Long input sequences often result in correspondingly lengthy decoded outputs, creating significant performance bottlenecks for LLM inference.

    \item \textbf{Training Misalignment:} Pre-trained LLMs, optimized for reasoning, coding, and mathematical tasks, are not designed for extracting and structuring content from HTML, presenting an opportunity for optimization.
\end{enumerate}

Our proposed solution is to fine-tune an existing SLM (under 2B parameters) to support longer contexts (ideally 512k) with two dedicated training objectives:

\textbf{Instructed Markdown Extraction}: This task aims to convert HTML documents into Markdown format based on specific instructions.
It focuses on extracting and converting relevant content from HTML documents into a structured Markdown representation, removing non-essential elements like navigation menus and advertisements.
While the default behavior targets conversion of a page's main content, users can provide specific instructions to customize the extraction scope and criteria, controlling how HTML content is transformed into structured Markdown while preserving meaning and document hierarchy. The extracted Markdown should accurately preserve the content based on user instructions while maintaining structure and organization.

\textbf{Schema-Guided JSON Extraction}: This task is schema-driven, requiring output to conform to a predefined JSON structure and set of fields.
The model needs to identify relevant data fields within the HTML document and map them to corresponding keys in the JSON schema.
The extracted representation must strictly adhere to the schema specifications, enabling downstream applications to utilize the data.

\section{Multi-Stage Data Synthesis and Training Pipeline}
\label{sec: pipeline}


We iteratively improve model performance by leveraging a curated dataset and synthetic data, refining the model through four stages: continued pretraining, supervised fine-tuning, direct preference optimization, and self-play iterative tuning.

\subsection{Datasets and Data Synthesis}
\label{sec: datasets}



Our training data comes from two sources: Our curated dataset, WebMarkdown-1M, consists of one million web pages converted to Markdown and JSON. Since no other existing datasets support our task, we pair this with synthetic data generated from our Draft-Refine-Critique process to produce high-quality training data.

\subsubsection{WebMarkdown-1M}

\begin{figure}[h]
    \centering
    \begin{subfigure}[b]{0.65\textwidth}
        \includegraphics[width=\textwidth]{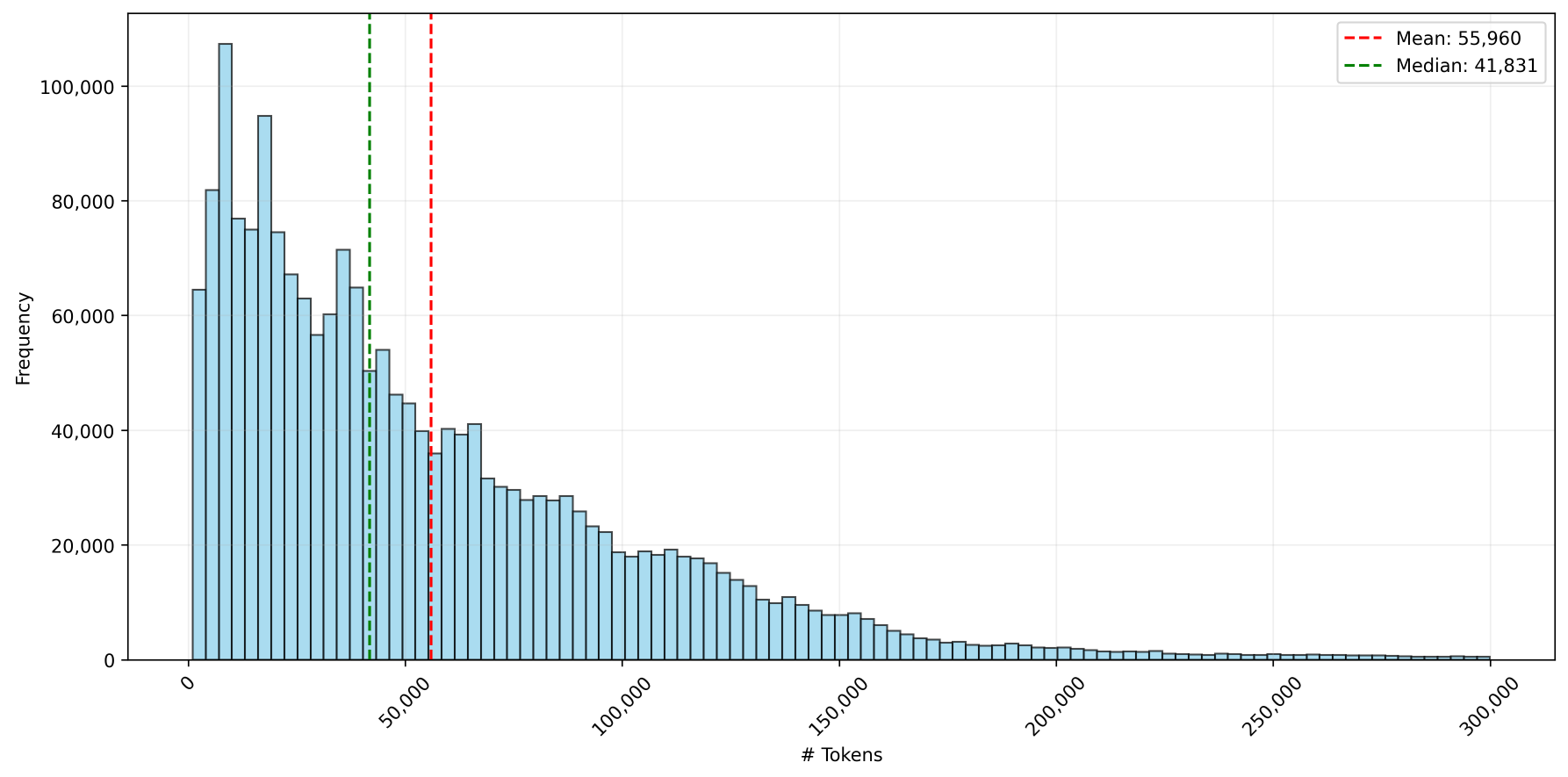}
        \caption{Token length distribution of HTML documents.}
        \label{fig:html-length-distribution}
    \end{subfigure}
    \hfill
    \begin{subfigure}[b]{0.34\textwidth}
        \includegraphics[width=\textwidth]{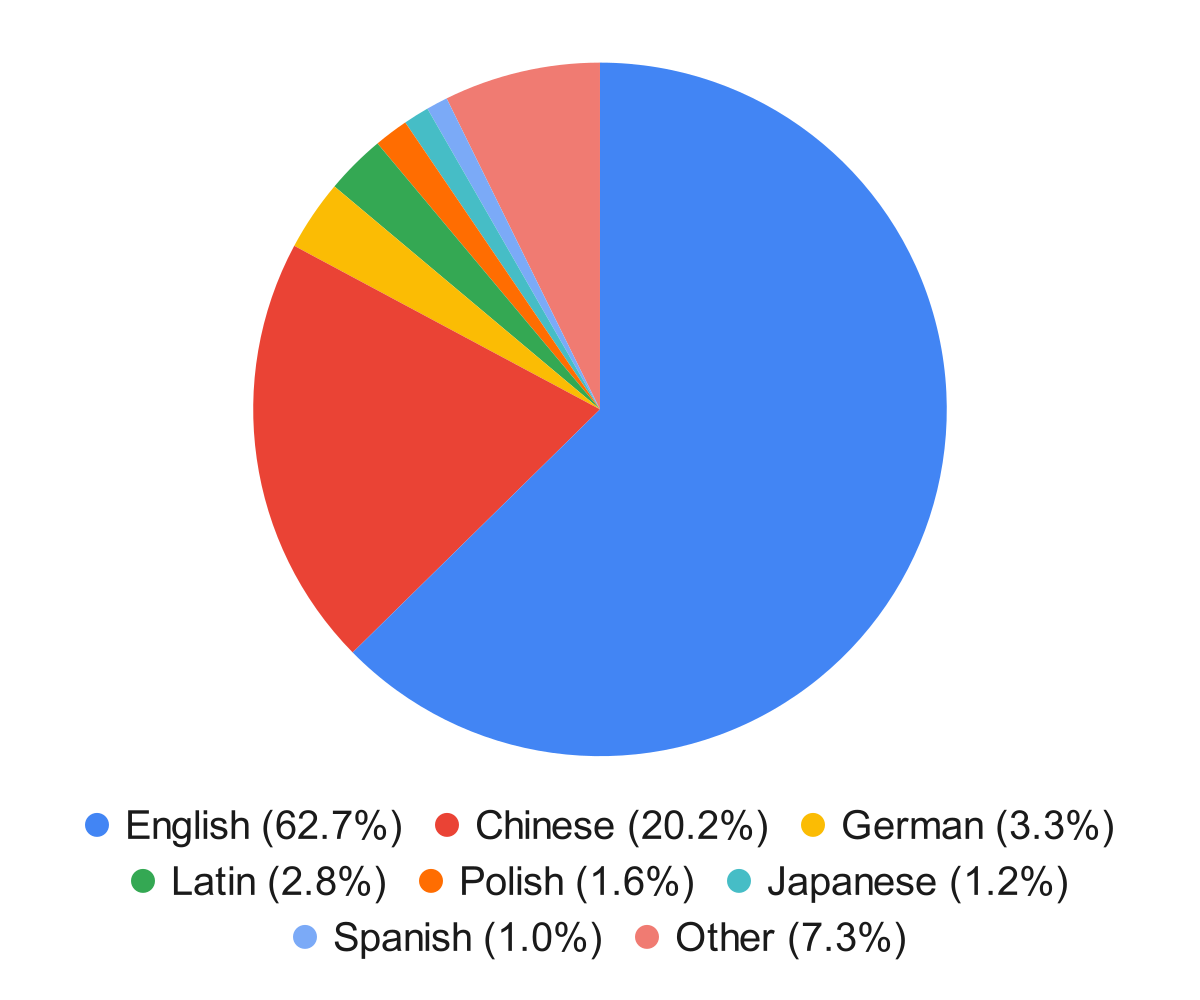}
        \caption{The language distribution of the HTML documents.}
        \label{fig:lang-distribution}
    \end{subfigure}
    \caption{Dataset statistics of \textbf{WebMarkdown-1M}.}
    \label{fig:dataset-stats}
\end{figure}

We construct a dataset named \textbf{WebMarkdown-1M} by randomly sampling one million URLs from the top 500 domains listed in the Common Crawl URL Index\footnote{\url{https://commoncrawl.org/blog/common-crawl-url-index}}.
Using a service called Reader\footnote{\url{https://jina.ai/reader}}, we convert the content of these URLs into Markdown and JSON formats.
Since the backbone model Qwen2.5 supports 29 global languages, we apply language detection and filtering to exclude documents in unsupported languages, resulting in the finalized \textbf{WebMarkdown-1M} dataset.
As shown in Figure \ref{fig:lang-distribution}, the dataset exhibits a marked language distribution, with English (62.7\%) and Chinese (20\%) dominating, collectively accounting for over 80\% of the content.
The document length analysis shown in Figure \ref{fig:html-length-distribution} reveals substantial variation, with documents averaging 56,000 tokens. This considerable length emphasizes the need for robust long-context processing capabilities to handle structured data extraction tasks.

\subsubsection{Synthetic Data with Draft-Refine-Critique}
\label{sec:synthetic_data_generation}

Due to the novelty of this task, no publicly-available datasets can be directly used for training models to perform HTML-to-Markdown conversion or schema-guided HTML-to-JSON extraction.
To address this, we employ synthetic data generation to create training data for Stage 2, Stage 3, and Stage 4 tuning.
Our method, \textbf{Draft-Refine-Critique}, is specifically designed to generate high-quality supervised tuning datasets and preference data.
Figure \ref{fig:data_synthetic_pipeline} illustrates the details of the pipeline:

\begin{figure}[htbp]
    \centering
    \includegraphics[width=0.95\columnwidth]{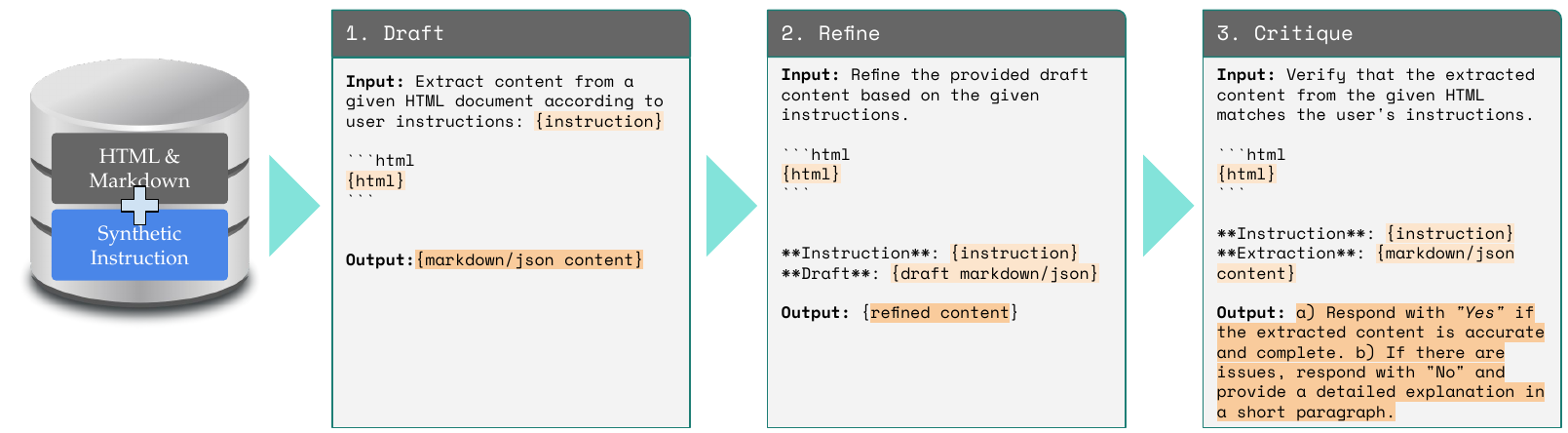}
    \caption{The \textbf{three-step} data synthesis pipeline for \shortname.}
    \label{fig:data_synthetic_pipeline}
\end{figure}

\begin{enumerate}
    \item \textbf{Draft:} Generate synthetic Markdown or JSON data based on specific instructions.
    While the generated samples align well with the desired output formats and content focus, they may include noise, redundancy, or inconsistencies.
    Despite these imperfections, this drafting step provides a broad and diverse set of training examples

    \item \textbf{Refine:} Refine the drafted data by removing redundancy, enforcing structural consistency, and ensuring adherence to format-specific conventions.
    An LLM-based review evaluates content accuracy, correctness, and real-world alignment, further improving data quality for subsequent stages.

    \item \textbf{Critique:} Employ a prompt-based evaluation to assess data accuracy and provide refinement feedback.
    By comparing model outputs against specified prompt requirements, this step detects discrepancies and inconsistencies, producing a binary judgment on whether the refined response meets quality standards, along with an explanation justifying the decision.
\end{enumerate}

The \textsc{Draft-Refine-Critique} process begins with raw HTML from the \textbf{WebMarkdown-1M} dataset.
Once the pipeline completes, we categorize the data into three distinct datasets based on the binary judgment outcome:

\begin{enumerate}
    \item \textbf{WebData-SFT-Filtered}: Includes only instances where the refined output successfully passed the Critique review, filtering out negative judgments.
    As a result, we obtain 250,000 high-quality instruction pairs for HTML-to-Markdown conversion and JSON schema-guided extraction.



    \item \textbf{WebData-SFT-Critique}: Contains 100,000 training examples from the Critique step.
        Each example pairs an input (the original HTML and its converted format from the Refine step) with an output (the critique's assessment and detailed explanation).
        We maintain a 1:2 ratio between examples with negative and positive critique assessments in the final dataset.

    \item \textbf{WebData-DPO-Preference}: Consists of preference pairs, where each sample includes an HTML input with its corresponding instruction, a desired output (validated by the Critique review), and an undesired output from the initial drafting stage.
    The final dataset comprises 150,000 preference triplets.
\end{enumerate}

The resulting \textbf{WebData-SFT-Filtered}, \textbf{WebData-SFT-Critique} and \textbf{WebData-DPO-Preference} datasets can therefore be utilized in the Stage 2 and Stage 3 tuning process.

\subsection{Training Pipeline}
\label{sec: training-pipeline}

We develop a multi-stage training pipeline through an iterative process, with each stage using a dedicated training dataset tailored as follows:

\textbf{Stage 1} begins with continued pretraining, where the model is warmed up using HTML-to-Markdown conversion data while extending its context length. In \textbf{Stage 2} consists of supervised fine-tuning (SFT) with synthetic data generated through our \textsc{Draft-Refine-Critique} pipeline. \textbf{Stage 3} builds on this by applying direct preference optimization (DPO) using data from the same pipeline, enabling the model to differentiate between high- and low-quality outputs. Finally, \textbf{Stage 4} introduces another round of SFT and DPO, utilizing data produced from the Stage 3 checkpoint in a process we call ``self-play iterative tuning.''

\subsubsection{Stage 1: Continued Pretraining}

We continue pre-training the base model \qwenbasemodel on the \textbf{WebMarkdown-1M} dataset.
The base model was originally pre-trained with a context length of 32,768 tokens and a RoPE base frequency of 1,000,000.
To improve our model's ability to handle longer sequences, we adopt the ring-zag attention mechanism~\citep{zigzagringattn}, extended context lengths and higher RoPE base frequencies.
Specifically, we implement a progressive context length expansion strategy during training, progressing through three stages of 32,768, 125,000, and ultimately 256,000 tokens, using a RoPE base frequency of 5,000,000.
Although the model is trained on sequences up to 256,000 tokens, it can extrapolate up to 512,000 during inference due to RoPE's inherent extrapolation capabilities.
To adapt to to increasing context lengths, we follow the approach outlined in \qwenname~\citep{qwen2024qwen25}, curating the training data to consist of 40\% sequences at the current maximum length and 60\% shorter sequences.
This balanced curriculum allows the model to incrementally adjust to longer contexts while preserving computational efficiency and maintaining training stability.


%

\subsubsection{Stage 2: Supervised Fine-tuning}

We apply supervised fine-tuning on top of the checkpoint produced from continued pretraining.
Instead of training a single model on all data types simultaneously, we train four separate specialized checkpoints, each focusing on different data types.
This includes two checkpoints trained with the \textbf{WebData-SFT-Filtered} dataset for HTML-to-Markdown and HTML-to-JSON tasks, and two checkpoints trained with the \textbf{WebData-SFT-Critique} dataset for the same tasks.
Recent research suggests that this approach can improve performance by mitigating data distribution conflicts~\citep{aakanksha2024mix}.

During model development, we encountered significant degeneration issues in the form of repetitive token generation. The model would either repeat individual tokens or enter loops of short token sequences until reaching the maximum output length. To address this, we incorporate contrastive loss~\citep{su2022contrastive} during training to encourage more discriminative and isotropic token representations. Our empirical testing show that this approach effectively reduced repetitive generation patterns.

We fine-tune each checkpoint for 25K steps with a learning rate of $4 \times 10^{-5}$, using a batch size of 1 and a maximum sequence length of 35K tokens. To combine the specialized capabilities of individual checkpoints into a single robust model, we apply linear parameter merging~\citep{pmlr-v162-wortsman22a} with weighted interpolation across the task-specific checkpoints.




\subsubsection{Stage 3: Direct Preference Optimization}


To further align the model with high-quality outputs, we employ Direct Preference Optimization (DPO)~\citep{rafailov2024direct}.
DPO is particularly effective for our use case as it allows the model to learn from paired comparisons of outputs, distinguishing between higher and lower quality renderings of the same HTML content.

We train the model using our \textbf{WebData-DPO-Preference} dataset, which leverages the paired outputs from our Draft-Refine-Critique pipeline described in Section~\ref{sec:synthetic_data_generation}.
Each HTML input yields a preference pair consisting of an initial draft and its refined version, providing natural quality differentials for DPO training without requiring manual annotations.
The DPO training uses a maximum sequence length of 65K tokens, a learning rate of $4 \times 10^{-6}$, and a batch size of 1.
This training phase effectively bridges the gap between basic task completion and expert-level performance, helping the model understand the subtle quality differences that distinguish refined outputs from initial drafts.


\subsubsection{Stage 4: Self-play Iterative Tuning}

To further enhance the model's performance and generalization, we introduce an additional training stage called \textbf{Self-play Iterative Tuning}.
This stage mirrors Stage 2 and Stage 3, which involve SFT and DPO, but with a key difference:
we apply the \textbf{Draft-Refine-Critique} process again, using the Stage 3 checkpoint instead of \qwenllm to generate draft data.
In this step, we regenerate the WebData-SFT-Filtered, WebData-SFT-Critique, and WebData-DPO-Preference datasets using the checkpoint obtained after weight merging and DPO in the previous stage.
We refer to this process as ``self-play" because the model generates instructed Markdown/JSON data and preference data independently, aiming to perfect the results.
Self-play iterative tuning creates a feedback loop where the model continuously generates better training data and improves its own performance through iterative training.

\section{Evaluation}
\label{sec:evaluation}

\subsection{Dataset}


We create a held-out test set from the \textbf{WebData-SFT-Filtered} dataset, carefully excluding all samples used in training.  

For the HTML to Markdown evaluation set, this is followed by a human verification process to ensure that HTML tags were accurately transformed into their corresponding Markdown syntax. For instance, heading tags such as \texttt{<h1>}, \texttt{<h2>}, and \texttt{<h3>} were correctly mapped to \texttt{\#}, \texttt{\#\#}, and \texttt{\#\#\#}, respectively. Additionally, we verify that the extracted content precisely matched the text enclosed within the original HTML tags and that no information was lost during the transformation.  
For the HTML to JSON evaluation set, we manually verify the correctness of JSON grammar using a grammar checker and ensured that the content of each JSON node accurately matched the corresponding HTML content without hallucination.  
Through this process, we produce a high-quality evaluation set containing 500 HTML to Markdown documents and 300 HTML to JSON documents.

\subsection{Instructed Markdown Extraction}

To evaluate the quality of the extracted Markdown, we focus on two key properties: \textbf{textual accuracy} and \textbf{structural preservation}.
Textual accuracy measures the content overlap between the extracted and reference markdown.
We quantify character-level edits using \textit{Levenshtein Distance}~\citep{levenshtein1966binary} and account for transpositions with \textit{Damerau-Levenshtein Distance}~\citep{damerau1964technique}, helping to detect typographical errors.
Structural preservation assesses how well the hierarchical organization and formatting of the Markdown are maintained.
We employ \textit{ROUGE-L}~\citep{lin2004rouge} to measure the longest matching subsequences, capturing both content matches and the preservation of Markdown formatting elements.
Additionally, \textit{Jaro-Winkler Similarity}~\citep{winkler1990string} is used to emphasize string similarity, particularly at the beginning of the text.

Together, these metrics provide a comprehensive evaluation of extraction quality across both textual content and Markdown structure.

We compare our model against two proprietary models: GPT-4o-2024-08-06 and Gemini 2.0 Flash, as well as a strong open-source baseline, Qwen2.5-32B-Instruct.
Additionally, we report results from three different training stages: supervised fine-tuning (Stage 2), direct preference tuning (Stage 3), and self-play iterative tuning (Stage 4).
This allows us to illustrate how each training stage contributes to the final performance improvements.
The results are presented in Table~\ref{tab:markdown-extraction-results}.

\begin{table}[ht]
\centering
\resizebox{\textwidth}{!}{
\begin{tabular}{>{\centering\arraybackslash}p{3cm} >{\centering\arraybackslash}p{3.5cm} cccc}
\toprule
\multirow{2}{*}{{Task}} & \multirow{2}{*}{{Model}} & \multicolumn{4}{c}{{Results}}  \\
\cmidrule{3-6}
&   & Rouge-L $\uparrow$ & Levenshtein $\downarrow$ & Damerau $\downarrow$ & Jaro-Winkler $\uparrow$ \\
\midrule
\multirow{6}{*}{\makecell[l]{Content\\Extraction}} & GPT-4o-2024-08-06               & 0.69 & 0.40 & 1283.54	& 0.75  \\
                                & Gemini 2.0 Flash              & 0.69 & 0.40	& 1341.14 &	0.74  \\
                                & Qwen2.5-32B-Instruct       & 0.71 & 0.41	& 1354.33 &	0.70 \\
\cdashline{2-6}
                                & \shortname-sft            & 0.71 & 0.35 & 1300.20 & 0.71  \\
                                & \shortname-dpo            & 0.84 & 0.22 & 1262.75	& 0.82  \\
                                & \shortname        & \textbf{0.86} & \textbf{0.20} & \textbf{928.15} & \textbf{0.83}  \\
\midrule
\multirow{6}{*}{\makecell[l]{Instructed\\Extraction}} & GPT-4o-2024-08-06                & 0.69 & 0.42 & \textbf{451.10} & 0.69  \\
                                & Gemini 2.0 Flash              & 0.64 & 0.45 & 766.62 & 0.70  \\
                                & Qwen2.5-32B-Instruct       & 0.68 & 0.43 & 501.50 & 0.69  \\
\cdashline{2-6}
                                & \shortname-sft                    & 0.68 & 0.41 & 927.34 & 0.74 \\
                                & \shortname-dpo                    & 0.70 & 0.38 &	673.62	& \textbf{0.75}  \\
                                & \shortname            & \textbf{0.72} & \textbf{0.37} & 748.10 & \textbf{0.75}  \\
\bottomrule
\end{tabular}
}
\caption{Results on two types of Markdown extraction tasks: (1) HTML main content conversion - transforming the primary content of web pages to Markdown format, and (2) instruction-guided extraction - converting specific HTML content to Markdown based on user instructions.}
\label{tab:markdown-extraction-results}
\end{table}

For the main Markdown extraction task, where models need to extract and convert the primary content from HTML to Markdown format, our results indicate significant improvements.
Compared to \textit{gpt-4o-2024-08-06}, \textit{gemini2-flash-expr} and \textit{qwen2.5-32B-instruct}, the final checkpoint of \shortname demonstrates a significant performance gain on all metrics.
For example, \shortname achieves a Rouge-L score of 0.86, which represents a 24.6\% increase compared to \textit{gpt-4o-2024-08-06} (0.69) and \textit{gemini2-flash-expr} (0.69), as well as a 21.1\% improvement over \textit{qwen2.5-32B-instruct} (0.71).
The direct preference optimization stage plays a pivotal role in the multi-stage training pipeline, contributing the most to the model’s final performance improvement.
After the direct preference optimization stage, we observe significant improvements across all metrics: Rouge-L increases from 0.71 to 0.84 and Jaro-Winkler from 0.71 to 0.82, while Levenshtein distance decreases from 0.35 to 0.22 and Damerau distance reduces from 1300.20 to 1262.75 (lower is better for distance metrics).
This highlights the effectiveness of the \textsc{Draft-Refine-Critique} synthetic data generation pipeline, where responses that pass critique are treated as preferred outputs, while drafted responses serve as undesired outputs.

We observe a similar trend in the instructed Markdown extraction task, where web content must be extracted according to user instructions.
Despite its relatively small size, \shortname outperforms all other competitors. However, its performance gains are less pronounced compared to main content extraction, highlighting the model’s size-related limitations, whereas larger models maintain consistent performance across both tasks.

\subsection{Schema Guided JSON Extraction}

To evaluate the quality of extracted JSON data, we convert both predicted and ground truth JSON objects into tree structures, where each node represents a key-value pair.
This transformation enables a systematic comparison of structural similarities and differences.  

We employ standard information retrieval metrics adapted to our tree-based representation:  
\textit{Precision} measures the proportion of correctly extracted nodes among all extracted nodes;  
\textit{Recall} indicates the proportion of ground truth nodes that are successfully extracted;  
\textit{F1-Score} provides a balanced measure as the harmonic mean of precision and recall.  
Additionally, we use \textit{Pass-Rate} to assess the proportion of outputs that are both syntactically-valid JSON and strictly adhere to the intended structure.

These metrics collectively provide a comprehensive evaluation of the quality and completeness of the extracted JSON data.
Evaluation results can be found in Table \ref{tab:Json extraction performence}.
\begin{table}[ht]
\centering
\begin{tabular}{ccccc} 
\toprule
\multirow{2}{*}{\centering Model} & \multicolumn{4}{c}{Results} \\
\cmidrule{2-5}
                       & F1   & Precision   & Recall   & Pass-Rate  \\
\midrule
GPT-4o-2024-08-06              &  \textbf{0.84}    &   0.84    &  \textbf{0.83}     &  \textbf{1.00}  \\
Gemini2-flash-expr          &   0.81   &   0.81    &   0.82    &  0.99  \\
Qwen2.5-32B-Instruct         &   0.83   &   \textbf{0.85}    &   \textbf{0.83}    &  \textbf{1.00}   \\
\cdashline{1-5}
\shortname-sft                 &   0.81   &   0.82    &   0.81    &  0.96   \\
\shortname-dpo                 &    0.82  &   0.82    &   0.82    &  0.97   \\
\shortname                 &    0.81  &   0.82    &   0.81    &   0.99  \\
\bottomrule
\end{tabular}
\caption{JSON extraction performance.}
\label{tab:Json extraction performence}
\end{table}

Unlike Markdown extraction, JSON extraction shows relatively stable performance across different training stages.
This suggests that the initial supervised fine-tuning stage effectively captures JSON's structured nature, while DPO primarily enhances reliability, as reflected in higher pass rates.
Despite having only 1.5 billion parameters, \shortname achieves solid performance on structured JSON extraction tasks, though it still lags behind larger models in this more complex task.

\section{Related Work}
\label{sec:related_work}





\subsection{HTML Understanding and Content Extraction}

Extracting structured HTML information has long been studied. Traditional methods use DOM trees (\cite{DBLP:journals/www/GuptaKGCS05}, \cite{DBLP:conf/www/GuptaKNG03}), template detection (\cite{DBLP:books/daglib/0022691}), or heuristic filtering (\cite{louvan2009extracting}).  

Recent advances leverage LLMs and advanced neural architectures. (\cite{DBLP:journals/corr/abs-2212-05238}) use GPT-3 for entity and relation extraction. (\cite{DBLP:conf/www/Wang0RFQL22}) introduce WebFormer, integrating web layout for structured extraction. (\cite{DBLP:conf/emnlp/GurNMSHCNFF23}) show LLMs pretrained on language tasks transfer well to web-based tasks. (\cite{DBLP:journals/corr/abs-2411-02959}) propose HtmlRAG, preserving HTML structure in RAG systems. 

These modern approaches highlight the growing role of LLMs and transformer-based architectures in structured information extraction, offering significant improvements over traditional heuristic-driven methods.

\subsection{Synthetic Data Generation}

Synthetic data generation has become essential in machine learning, addressing challenges in cost, availability, and accessibility of real-world labeled data. By replicating real data characteristics, it supports robust model development across applications.

In NLP, synthetic data has proven valuable for enhancing training datasets in tasks such as machine translation~\citep{sennrich2016improving}, text summarization~\citep{dong2019unified}, and question answering~\citep{alberti2019synthetic}, leading to improved model performance and generalization.

The field employs multiple approaches, including data augmentation techniques like paraphrasing, back-translation, and noise injection~\citep{feng2021survey}. Additionally, pre-trained language models such as GPT-3 can generate structured synthetic text data, producing content that adheres to specific patterns or structural requirements~\citep{brown2020language}.

While synthetic data offers scalability and customization, maintaining data quality and relevance is critical. Our approach incorporates validation procedures and quality controls to ensure the synthetic data effectively serves its purpose in improving language model capabilities, balancing its benefits with the risks of noise and bias.

\section{Conclusion}


This paper introduces \shortname, a compact yet powerful language model for structured data extraction from long-context documents.
Using our novel three-stage data synthesis pipeline and comprehensive training strategy, we demonstrate that small models can handle complex tasks traditionally reserved for larger models.
While larger models still lead in some complex tasks, \shortname shows competitive performance in HTML-to-Markdown conversion, particularly for main content extraction, offering a more resource-efficient solution for real-world applications.
This work opens new possibilities for efficient, scalable structured data extraction systems, with future directions focusing on extending capabilities to other formats and supporting more modalities while maintaining computational efficiency.

\clearpage

\bibliography{iclr2025_conference}
\bibliographystyle{iclr2025_conference}


\end{document}